\documentclass[conference,10pt,letterpaper]{IEEEtran}
\IEEEoverridecommandlockouts

\usepackage{cite}
\usepackage{amsmath,amssymb,amsfonts}
\usepackage{graphicx}
\usepackage{textcomp}
\usepackage{xcolor}
\usepackage{multirow}
\usepackage{url}
\usepackage{orcidlink}
\usepackage{microtype}
\usepackage{stfloats}

\def\BibTeX{{\rm B\kern-.05em{\sc i\kern-.025em b}\kern-.08em
    T\kern-.1667em\lower.7ex\hbox{E}\kern-.125emX}}

\begin{document}
\bstctlcite{IEEEexample:BSTcontrol}

\title{Smart Insole Human Activity Recognition for Continuous Monitoring in Elderly Care}

\newcommand{\conferenceauthorcell}[5]{%
\begin{minipage}[t]{0.43\textwidth}
\centering
{\normalfont\sublargesize #1}\par\vspace{0.35em}
{\normalfont\normalsize\textit{#2}}\par
{\normalfont\normalsize\textit{#3}}\par
{\normalfont\normalsize #4}\par
{\normalfont\normalsize #5}
\end{minipage}}

\author{\begin{minipage}{\textwidth}
\centering
\begin{tabular}{@{}c@{\hspace{0.07\textwidth}}c@{}}
\conferenceauthorcell{Edwin Rios~\orcidlink{0000-0002-8133-5819}}
{Department of Electrical and Computer Engineering}
{Worcester Polytechnic Institute}
{Worcester, MA, USA}
{earios1@wpi.edu}
&
\conferenceauthorcell{Antony Garcia~\orcidlink{0000-0002-5795-2104}}
{Facultad de Ingenier\'{i}a El\'{e}ctrica}
{Universidad Tecnol\'{o}gica de Panam\'{a}}
{Panama City, Panama}
{antony.garcia@utp.ac.pa}
\end{tabular}\par\vspace{1.45em}
\begin{tabular}{@{}c@{\hspace{0.07\textwidth}}c@{}}
\conferenceauthorcell{Fengpei Yuan~\orcidlink{0000-0002-5653-5552}}
{Department of Robotics Engineering}
{Worcester Polytechnic Institute}
{Worcester, MA, USA}
{fyuan3@wpi.edu}
&
\conferenceauthorcell{Xinming Huang~\orcidlink{0000-0003-0584-3448}}
{Department of Electrical and Computer Engineering}
{Worcester Polytechnic Institute}
{Worcester, MA, USA}
{xhuang@wpi.edu}
\end{tabular}
\end{minipage}%
}

\maketitle

\begin{abstract}
Falls in older adults are often preceded by changes in mobility, balance, and postural transitions. This paper presents a wireless smart insole platform and machine-learning workflow for recognizing sitting, standing, walking, and unstable walking from plantar-pressure and inertial signals. Each insole integrates 16 active pressure-sensing locations and a six-dimensional IMU stream consisting of tri-axial acceleration and angular velocity. Data were collected from 15 healthy adults at 80~Hz and segmented into overlapping windows. Window length and candidate model families were first screened with stratified 10-fold cross-validation; the primary performance estimate was then obtained with participant-independent 5-fold Stratified Group cross-validation, ensuring that all windows from a participant remained in a single fold. Under this protocol, Histogram-Based Gradient Boosting (HGB) achieved macro-F1 scores of 0.954 and 0.959 for the left and right feet, respectively, and 0.980 with bilateral sensing. A compact 1D-CNN evaluated with the same participant-independent folds did not significantly outperform HGB ($p=0.0625$). The results show that low-profile footwear sensing can infer activity state from pressure and IMU measurements for participants unseen during training, establishing a basis for activity monitoring and fall prevention in elderly care.
\end{abstract}

\section{Introduction}
Falls are a major threat to independence in later life and are associated with injury, loss of confidence, institutionalization, and high health-care use. Gait and balance deficits are among the most important fall-risk factors in older adults~\cite{rubenstein_falls_2006}, and the risk is increased in people with cognitive impairment or dementia, where unsafe ambulation and reduced hazard awareness complicate supervision~\cite{fernando_risk_2017,modarresi_gait_2019}. For elder-care facilities, prevention requires more than detecting an impact after a fall has occurred. A useful system should identify the mobility states that often precede risk, including sitting, standing, gait initiation, and unstable walking.

Postural transitions are central to this problem. Sit-to-stand requires coordinated forward trunk motion, lower-limb force generation, foot loading, and balance recovery before walking begins. Instrumented transition studies have shown that stand-sit and sit-stand characteristics can be measured using wearable sensors and are relevant to fall-risk evaluation in older adults~\cite{najafi_measurement_2002}. Sit-to-stand transitions extracted during activities of daily living have also been shown to reveal acute fall risk~\cite{pozaic_sit_2016}. In practical elder care, a high-risk episode may begin when a resident rises from a chair and starts walking without guidance. Detecting that sequence early could allow a caregiver notification, a check-in, or a guidance cue before a fall occurs. The same state stream can also support longitudinal review. For example, changes in the frequency of chair-rise episodes, time spent standing, or the number of short walking bouts may indicate decline, agitation, or increased need for assistance. Unlike camera-based monitoring, footwear sensing can preserve privacy in bedrooms and bathrooms while still measuring the mobility events most relevant to falls.

Wearable human activity recognition (HAR) provides a route toward this type of monitoring. Body-worn inertial sensors have been widely studied for classifying activities of daily living~\cite{lara_survey_2013}; however, wrist, trunk, and thigh devices can be forgotten, removed, or poorly tolerated during continuous monitoring. Footwear is a natural sensing location for mobility because plantar loading and foot motion directly encode stance, gait, and weight transfer. Smart insoles can therefore capture the mechanical events most relevant to walking and balance while remaining unobtrusive inside shoes. From a systems perspective, in-shoe sensing is also attractive because it can be paired with daily footwear rather than requiring a separate wearable to be charged, placed, and worn correctly~\cite{piau_smart_2015}. This matters in dementia care, where adherence to conventional wearable devices can be difficult and where staff must minimize additional handling burden.

This work evaluates a custom smart insole as an elder-care HAR platform. The paper does not claim to predict future falls directly. Instead, it tests the prerequisite sensing question: can sitting, standing, walking, and balance-challenging gait be reliably inferred from low-profile pressure and IMU sensors embedded in footwear? The contributions are: (i) a compact insole integrating 16 active plantar-pressure sensing locations and six IMU dimensions per foot; (ii) a signal-processing and classification workflow for unilateral and bilateral activity recognition; (iii) a direct unilateral--bilateral comparison whose main performance estimate is confirmed with participant-independent validation; and (iv) a comparison of a classical ensemble with a compact 1D-CNN. 

\section{Background and Related Work}
\subsection{Fall-risk monitoring and postural transitions}
Clinical fall prevention depends on identifying risk factors early enough to intervene. Reviews of older-adult fall risk consistently identify gait impairment, balance deficits, weakness, cognitive impairment, and environmental factors as important contributors~\cite{rubenstein_falls_2006,fernando_risk_2017,modarresi_gait_2019}. Recent reviews conclude that wearable sensors can provide objective fall-risk information in older adults, including those with cognitive impairment, but that performance depends strongly on sensor placement, features, and evaluation protocol~\cite{bezold_sensor-based_2021}.

Among daily movements, sit-to-stand and stand-to-sit transitions are especially informative because they combine strength, coordination, anticipatory postural adjustment, and balance recovery. Najafi et al. quantified postural transitions with a miniature gyroscope and applied transition parameters to fall-risk evaluation~\cite{najafi_measurement_2002}. Pozaic et al. showed that sit-to-stand features recorded during daily activity could distinguish older fallers from non-fallers~\cite{pozaic_sit_2016}. These studies motivate continuous detection of the activity states surrounding chair-rise and gait initiation. They also show why post-fall alarms are not sufficient for prevention: once impact has occurred, the opportunity to guide or assist the person has been missed. A sensing system for care settings should therefore identify the pre-fall context, including seated rest, transition to standing, and the onset of walking. The present study focuses on these foundational activity states, which can later be combined into higher-level transition detectors.

\subsection{Smart insoles for activity recognition}
Insole-based systems complement trunk- or wrist-worn sensors by measuring both body motion and foot-ground interaction~\cite{sazonov_monitoring_2011,ngueleu_validity_2019}. smart insole HAR studies show that pressure and inertial signals can classify activities such as sitting, standing, walking, stairs, and sit-to-stand~\cite{sazonov_monitoring_2011,darco_assessing_2022,darco_deephar_2023,ngueleu_validity_2019}. Pressure sensors capture where and when the foot is loaded, which helps separate sitting from standing and reflects weight transfer. The IMU captures acceleration and angular velocity, which become prominent during gait initiation and walking. Combining these modalities in footwear therefore matches the mechanics of the elder-care states of interest. Prior smart insole studies also suggest that the number and placement of pressure sensors affect recognition performance. Dense commercial pressure arrays can provide detailed maps, but they increase cost, wiring complexity, data rate, and power demand~\cite{lin_smart_2016}. A lower-count sensor layout may be more realistic for a device that must fit in ordinary shoes and run for long periods. The 16-point array used here is intended as a practical compromise between coverage of heel, midfoot, and forefoot loading and hardware simplicity.

A second design question is whether one insole is sufficient or whether both feet should be instrumented. Unilateral sensing is attractive because it reduces device cost and charging burden, but bilateral sensing can represent left-right load transfer, stance asymmetry, and the coordination of the two feet during gait. These signals may be especially important in transition-related monitoring, where the first step after standing and the redistribution of weight between the two feet can reveal whether a person is preparing to walk. The bilateral comparison in this paper therefore addresses a practical deployment question rather than a purely algorithmic one.

\subsection{Embedded and edge-compatible inference}
A fall-prevention monitor should operate close to the body to reduce latency and dependence on external infrastructure. Modern microcontrollers support lightweight edge inference, but model choice must balance accuracy, memory, inference time, and energy use. Classical tree ensembles can be competitive for flattened sensor features, while convolutional models can improve performance when temporal structure is important. In this paper, both options are evaluated: Histogram-Based Gradient Boosting is used as a strong classical baseline, and a compact 1D-CNN is used to test whether temporal patterns in pressure and IMU windows improve recognition.

\section{Methods}
\subsection{smart insole platform}
The prototype consists of a flexible insole instrumented with 16 active resistive pressure-sensing locations and an IMU providing tri-axial acceleration and tri-axial angular velocity. For spatial representation, the pressure layout is stored as 18 pressure channels, with two fixed zero-valued positions used to preserve the matrix geometry; together with the six IMU channels, this yields the 24-channel unilateral input used by the 1D-CNN. Pressure and IMU signals were sampled at 80~Hz. The sensing layer was embedded in a shoe-compatible form factor and connected to a compact wireless electronics module for data acquisition and streaming. Sensor data were transmitted to a laptop over Wi-Fi and stored as labeled CSV files for offline model development~\cite{edwin_rios_antony_garcia_2026}. For bilateral recordings, a host computer received both insole streams and periodically transmitted synchronization beacons to the two devices so that corresponding left- and right-foot data could be combined into paired windows. The current study evaluates the sensing and modeling workflow offline.

\begin{figure}[t]
    \centering
    \includegraphics[width=0.92\linewidth]{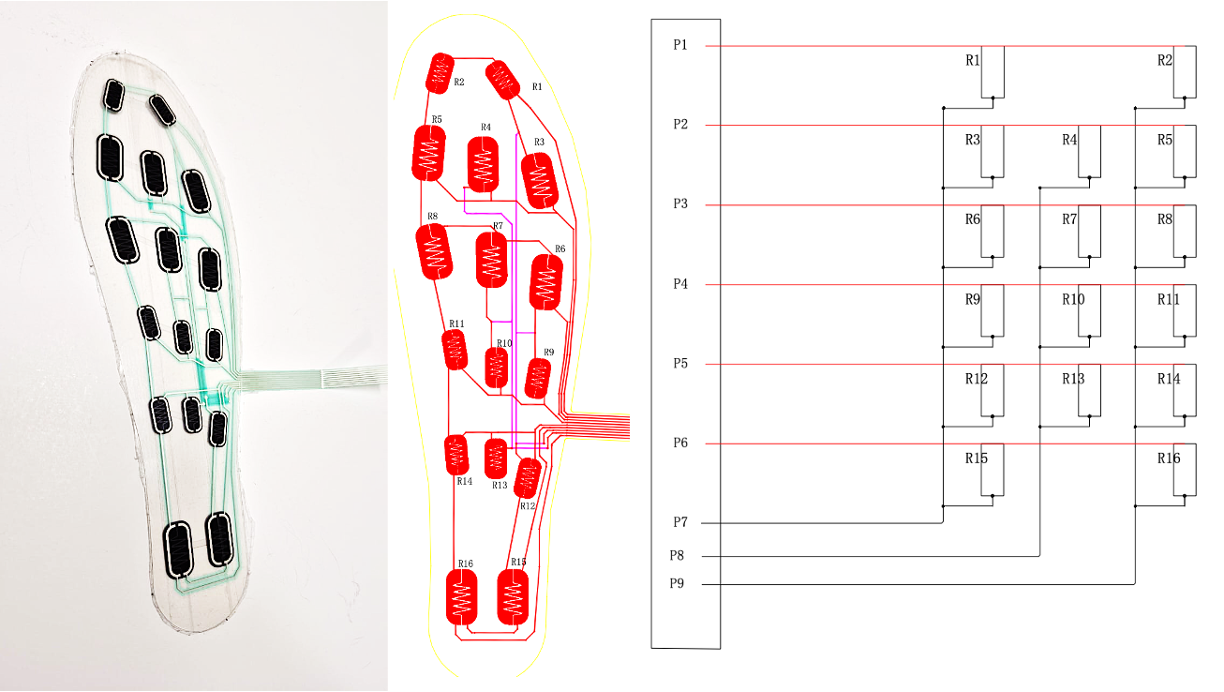}
    \caption{Prototype smart insole with embedded plantar-pressure sensing and inertial measurement electronics.}
    \label{fig:insole}
\end{figure}

\begin{figure}[t]
    \centering
    \includegraphics[width=0.92\linewidth]{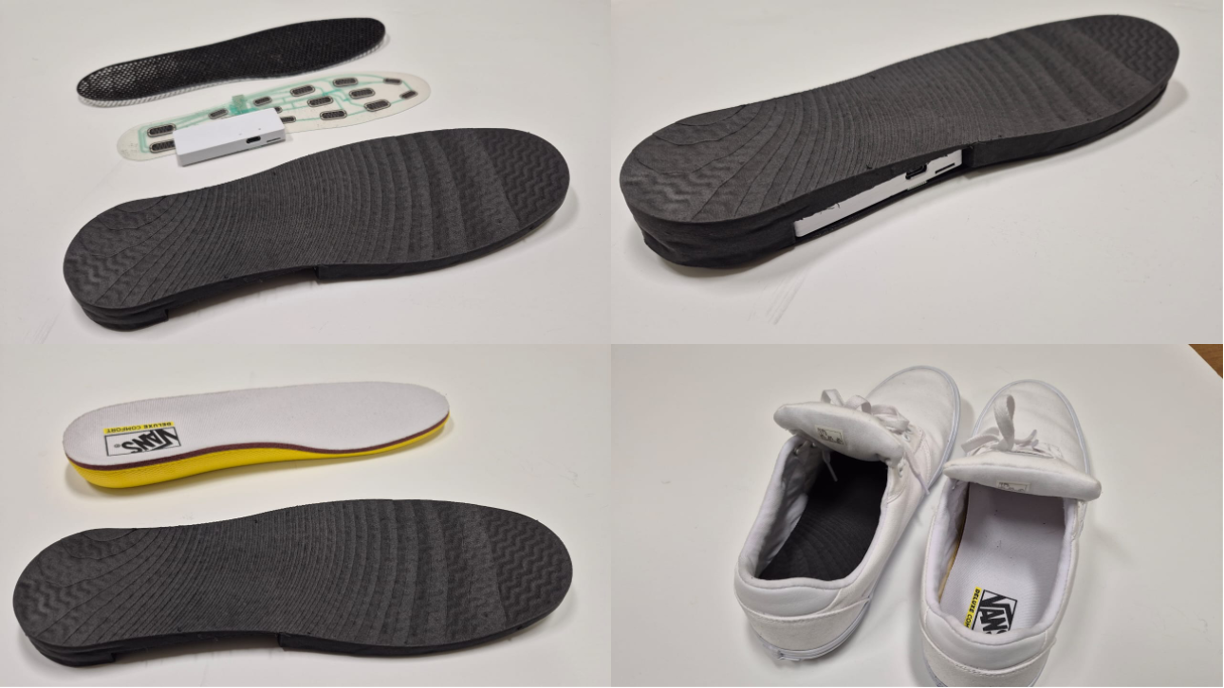}
    \caption{\textbf{Layered construction of the insole.} The picture shows the printed structural layers, embedded electronics module, pressure-sensing array, and final assembly inside a commercial shoe. The three-layer design accommodates the sensor matrix, cabling, and enclosure while preserving comfort and flexibility.}
    \label{fig:sandwich}
\end{figure}

\subsection{Participants and protocol}
Data were collected from 15 healthy adult participants as an initial baseline before extending the system to older or clinical populations. The target application is elder-care fall prevention, but healthy participants were used in this phase to validate the sensing and analysis workflow under controlled conditions. All recordings were performed indoors. Participants completed walking, standing, sitting, and tandem walking tasks, each lasting five minutes. Tandem walking was included as a balance-challenging gait condition. The protocol and participant characteristics are shown in Tables~\ref{tab:protocol} and~\ref{tab:demographics}.

\begin{figure*}[t]
    \centering
    \includegraphics[width=0.90\textwidth]{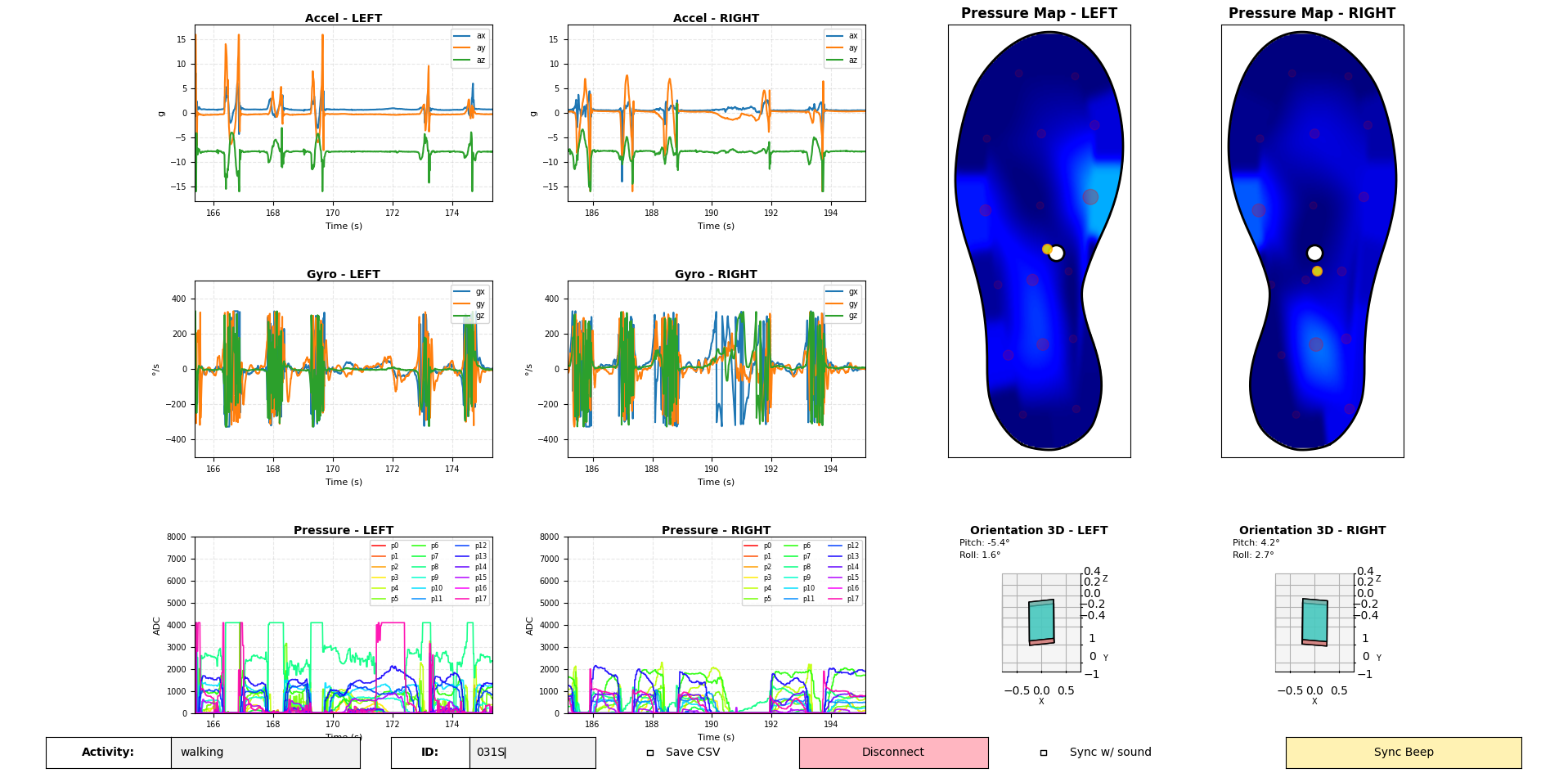}
    \caption{\textbf{Real-time data visualization and control dashboard.} This interface shows live sensor data, activity classification results, and system status. It also lets users save data to disk and label signals, making data collection and labeling easier during experiments.}
    \label{fig:dashboard}
\end{figure*}

\begin{table}[!t]
\centering
\caption{Summary of data collection protocol. The table lists all activities performed during the data collection session, including baseline postures, gait tasks, and the tandem-walking balance challenge, as well as their duration.}
\label{tab:protocol}
\begin{tabular}{l c}
\hline
\textbf{Activity} & \textbf{Duration} \\
\hline
Walking (normal gait) & 5~min \\
Standing (natural posture) & 5~min \\
Sitting (natural posture) & 5~min \\
Tandem walking & 5~min \\
\hline
\end{tabular}
\end{table}

\begin{table}[!t]
\centering
\caption{Participant demographic summary. The table provides an overview of the demographic characteristics of the 15 healthy adult participants included in the dataset, including gender distribution, age range, height range, shoe size range, and health status relevant to gait analysis.}
\label{tab:demographics}
\begin{tabular}{l c}
\hline
\textbf{Characteristic} & \textbf{Value} \\
\hline
Total Participants & 15 \\
Female & 6 \\
Male & 9 \\
Age Range & 18--54 years \\
Most Represented Group & 25--34 years (53\%) \\
Height Range & 160--182 cm \\
US Shoe Size Range & 7--11.5 \\
Health Status & No reported gait impairments \\
\hline
\end{tabular}
\end{table}

\subsection{Segmentation and classical machine learning}
Signals were segmented into windows ranging from 40 to 240 samples with 25\% overlap. A Random Forest classifier was first used to select the window size. Separate left- and right-foot datasets were evaluated with stratified 10-fold cross-validation, using accuracy, weighted precision, weighted recall, and weighted F1-score. The selected window length was 160 samples, which provided the best overall trade-off between temporal context and classification performance (Table~\ref{tab:performance_metrics}). In the intended monitoring application, this window length is also short enough to support near-real-time state updates while retaining enough samples to capture stance and gait-cycle structure. Shorter windows may react faster but can miss a full loading pattern, whereas longer windows can delay detection of a chair-rise or gait-initiation event.

\begin{table*}[!t]
\centering
\caption{Performance metrics for left and right feet using different sample window sizes. The metrics include accuracy, precision, recall, and F1-score, with confidence intervals computed using the t-distribution. These 10-fold window-level results are used for window-size screening only; participant-independent results are reported separately.}
\label{tab:performance_metrics}

\resizebox{\textwidth}{!}{%
\begin{tabular}{llllllll}
\hline
\multirow{2}{*}{Feet}  & \multirow{2}{*}{Metrics (95\% CI)} & \multicolumn{6}{c}{Samples per Window}                                                                                                                  \\ \cline{3-8} 
                       &                                    & \multicolumn{1}{c}{40} & \multicolumn{1}{c}{80} & \multicolumn{1}{c}{120} & \multicolumn{1}{c}{160} & \multicolumn{1}{c}{200} & \multicolumn{1}{c}{240} \\ \hline
\multirow{4}{*}{Left}  & Accuracy                           & 0.955 [0.953, 0.958]   & 0.959 [0.955, 0.963]   & 0.961 [0.956, 0.966]    & 0.965 [0.962, 0.969]    & 0.958 [0.951, 0.964]    & 0.962 [0.957, 0.967]    \\ 
                       & Precision                          & 0.956 [0.954, 0.959]   & 0.960 [0.956, 0.963]   & 0.962 [0.957, 0.967]    & 0.966 [0.962, 0.969]    & 0.958 [0.952, 0.965]    & 0.962 [0.957, 0.968]    \\ 
                       & Recall                             & 0.955 [0.953, 0.958]   & 0.959 [0.955, 0.963]   & 0.961 [0.956, 0.966]    & 0.965 [0.962, 0.969]    & 0.958 [0.951, 0.964]    & 0.962 [0.957, 0.967]    \\ 
                       & F1-Score                           & 0.956 [0.953, 0.958]   & 0.959 [0.955, 0.963]   & 0.961 [0.956, 0.966]    & 0.965 [0.962, 0.969]    & 0.958 [0.952, 0.964]    & 0.962 [0.957, 0.967]    \\ \hline
\multirow{4}{*}{Right} & Accuracy                           & 0.955 [0.952, 0.957]   & 0.961 [0.957, 0.965]   & 0.967 [0.964, 0.971]    & 0.968 [0.963, 0.973]    & 0.966 [0.958, 0.973]    & 0.965 [0.957, 0.974]    \\ 
                       & Precision                          & 0.956 [0.953, 0.958]   & 0.961 [0.958, 0.965]   & 0.968 [0.964, 0.971]    & 0.968 [0.963, 0.973]    & 0.966 [0.959, 0.973]    & 0.966 [0.957, 0.974]    \\ 
                       & Recall                             & 0.955 [0.952, 0.957]   & 0.961 [0.957, 0.965]   & 0.967 [0.964, 0.971]    & 0.968 [0.963, 0.973]    & 0.966 [0.958, 0.973]    & 0.965 [0.957, 0.974]    \\ 
                       & F1-Score                           & 0.955 [0.952, 0.958]   & 0.961 [0.957, 0.965]   & 0.967 [0.964, 0.971]    & 0.968 [0.963, 0.973]    & 0.966 [0.958, 0.973]    & 0.965 [0.957, 0.974]    \\ \hline
\end{tabular}%
}

\end{table*}

Using the selected 160-sample window, 21,069 samples were obtained for the four activity classes. Class counts were approximately balanced, reducing the likelihood that high accuracy was driven by a dominant activity label (Table~\ref{tab:class_distribution}).

\begin{table}[!t]
\centering
\caption{Distribution of collected samples by activity class and foot, showing the total count and percentage for each class, along with the corresponding left and right foot contributions.}
\label{tab:class_distribution}
\begin{tabular}{lccc}
\hline
\multicolumn{1}{c}{\textbf{Class}} & \textbf{Count}  & \textbf{Left Foot} & \textbf{Right Foot} \\ \hline
Sitting                            & 5,066 (24.0\%) & 2,368 (46.7\%)     & 2,698 (53.3\%)      \\
Standing                           & 5,559 (26.4\%) & 3,257 (58.6\%)     & 2,302 (41.4\%)      \\
Tandem                             & 5,043 (23.9\%) & 2,629 (52.1\%)     & 2,414 (47.9\%)      \\
Walking                            & 5,401 (25.6\%) & 2,774 (51.4\%)     & 2,627 (48.6\%)      \\ \hline
\multicolumn{1}{c}{\textbf{Total}} & 21,069 (100\%)  & 10,028 (47.6\%)    & 11,041 (52.4\%)     \\ \hline
\end{tabular}
\end{table}

Classical classifiers were then evaluated using flattened pressure and IMU features. The model-search stage used default Scikit-learn estimator settings as a baseline screen rather than aggressively tuning one model family. For paired comparisons, the same fold partitions were reused for each model variant. Confidence intervals were computed from fold-level metrics as
\begin{equation}
\bar{x} \pm t_{0.975,\,K-1}\frac{s}{\sqrt{K}}
\end{equation}
where $K=10$ folds ($t_{0.975,9}=2.262$), $\bar{x}$ is the fold mean, and $s$ is the fold-level sample standard deviation. Paired Wilcoxon signed-rank tests were used for unilateral--bilateral and HGB--CNN comparisons because the same folds were used for each paired condition~\cite{rainio_evaluation_2024}. Effect size was reported as $r=Z/\sqrt{N}$.

\subsection{Participant-independent validation}
To evaluate how well the models generalize to new participants, we used 5-fold Stratified Group cross-validation. Participant identity was used to define the groups, ensuring that all windows from the same participant were kept together in either the training or test set and never split between both. The same folds were used for all unilateral, bilateral, HGB, and CNN comparisons to ensure a fair evaluation. These participant-independent results are used as the main performance results in this paper, while the original 10-fold window-level cross-validation is used only for window-size and model selection. For the HGB--CNN comparison, statistical significance was evaluated using the Wilcoxon signed-rank test, and effect size was reported as $r = Z/\sqrt{N}$, consistent with the original analysis.

\subsection{1D-CNN reference model}
A 1D-CNN was implemented to classify activity directly from multichannel time-series windows. The model contains three temporal convolutional blocks followed by adaptive average pooling and a small fully connected classifier. It was trained with cross-entropy loss and AdamW (learning rate $10^{-3}$, weight decay $10^{-4}$), batch size 128, and dropout of 0.25. Per-channel z-score normalization was fit within each training fold and applied to validation/test splits. The 24 unilateral input channels consist of an 18-position pressure representation (16 active sensors and two fixed zero positions that preserve the spatial layout) plus six IMU channels. The architecture is shown in Table~\ref{tab:cnn_arch}.

\begin{table}[!t]
\centering
\caption{Architecture of the proposed 1D convolutional neural network for unilateral input, showing layer configuration and corresponding output dimensions at each stage.}
\label{tab:cnn_arch}
\resizebox{\columnwidth}{!}{%
\begin{tabular}{clc}
\hline
\textbf{Stage}           & \multicolumn{1}{c}{\textbf{Layer}}                     & \textbf{Output (unilateral)} \\ \hline
Input                    & --                                                     & $24 \times 160$              \\ \hline
\multirow{2}{*}{Block 1} & Conv1D(24$\rightarrow$32, $k{=}7$, pad 3) + BN + ReLU  & $32 \times 160$              \\
                         & MaxPool($k{=}2$)                                       & $32 \times 80$               \\ \hline
\multirow{2}{*}{Block 2} & Conv1D(32$\rightarrow$64, $k{=}7$, pad 3) + BN + ReLU  & $64 \times 80$               \\
                         & MaxPool($k{=}2$)                                       & $64 \times 40$               \\ \hline
\multirow{2}{*}{Block 3} & Conv1D(64$\rightarrow$128, $k{=}5$, pad 2) + BN + ReLU & $128 \times 40$              \\
                         & MaxPool($k{=}2$)                                       & $128 \times 20$              \\ \hline
Pooling                  & AdaptiveAvgPool($1$)                                   & $128 \times 1$               \\ \hline
Classification Head      & Linear(128$\rightarrow$64) + ReLU + Dropout(0.25)      & 64                           \\ \hline
Output                   & Linear(64$\rightarrow K$)                              & $K$                          \\ \hline
\end{tabular}%
} 
\end{table}

\section{Results}
Table~\ref{tab:model_selection_metrics} shows the original window-level model-selection results. HGB achieved the strongest classical performance, with 0.970 accuracy for the left foot and 0.968 for the right foot, and an offline inference time of 0.007~s. These timing values compare software models only and are not measurements of embedded end-to-end latency.

\begin{table*}[!t]
\centering
\caption{Performance metrics and processing times for various classifiers applied to left and right foot data.}
\label{tab:model_selection_metrics}
\resizebox{\textwidth}{!}{%
\begin{tabular}{llcccccc}
\hline
\multirow{2}{*}{\textbf{Foot}} & \multicolumn{1}{c}{\multirow{2}{*}{\textbf{Model}}} & \multicolumn{4}{c}{\textbf{Mean performance metrics {[}95\% Confidence Interval{]}}}                      & \multicolumn{2}{c}{\textbf{Processing time (s)}} \\ \cline{3-8} 
                               & \multicolumn{1}{c}{}                                & \textbf{Accuracy}        & \textbf{Precision}       & \textbf{Recall}          & \textbf{F1-Score}        & \textbf{Training}      & \textbf{Inference}      \\ \hline
\multirow{10}{*}{Left}        & HistGradientBoostingClassifier                      & 0.970 {[}0.966, 0.973{]} & 0.970 {[}0.966, 0.974{]} & 0.970 {[}0.966, 0.973{]} & 0.970 {[}0.966, 0.973{]} & 182.810                & 0.007                   \\
                               & GradientBoostingClassifier                          & 0.963 {[}0.958, 0.967{]} & 0.963 {[}0.959, 0.967{]} & 0.963 {[}0.958, 0.967{]} & 0.963 {[}0.959, 0.967{]} & 10840.420              & 0.014                   \\
                               & ExtraTreesClassifier                                & 0.961 {[}0.957, 0.965{]} & 0.962 {[}0.958, 0.966{]} & 0.961 {[}0.957, 0.965{]} & 0.961 {[}0.957, 0.966{]} & 35.110                 & 0.012                   \\
                               & RandomForestClassifier                              & 0.959 {[}0.955, 0.963{]} & 0.960 {[}0.956, 0.964{]} & 0.959 {[}0.955, 0.963{]} & 0.959 {[}0.955, 0.964{]} & 111.570                & 0.010                   \\
                               & SVC                                                 & 0.959 {[}0.954, 0.964{]} & 0.959 {[}0.955, 0.964{]} & 0.959 {[}0.954, 0.964{]} & 0.959 {[}0.954, 0.964{]} & 155.520                & 5.158                   \\
                               & MLPClassifier                                       & 0.959 {[}0.956, 0.961{]} & 0.959 {[}0.957, 0.961{]} & 0.959 {[}0.956, 0.961{]} & 0.959 {[}0.956, 0.961{]} & 412.550                & 0.004                   \\
                               & KNeighborsClassifier                                & 0.901 {[}0.896, 0.907{]} & 0.910 {[}0.905, 0.914{]} & 0.901 {[}0.896, 0.907{]} & 0.900 {[}0.894, 0.905{]} & 0.710                  & 0.120                   \\
                               & ExtraTreeClassifier                                 & 0.876 {[}0.868, 0.885{]} & 0.876 {[}0.868, 0.884{]} & 0.876 {[}0.868, 0.885{]} & 0.876 {[}0.867, 0.884{]} & 0.490                  & 0.002                   \\
                               & DecisionTreeClassifier                              & 0.860 {[}0.851, 0.870{]} & 0.860 {[}0.850, 0.870{]} & 0.860 {[}0.851, 0.870{]} & 0.860 {[}0.850, 0.870{]} & 110.980                & 0.002                   \\
                               & AdaBoostClassifier                                  & 0.776 {[}0.757, 0.796{]} & 0.782 {[}0.764, 0.800{]} & 0.776 {[}0.757, 0.796{]} & 0.777 {[}0.757, 0.796{]} & 503.560                & 0.078                   \\ \hline
\multirow{10}{*}{Right}        & HistGradientBoostingClassifier                      & 0.968 {[}0.966, 0.971{]} & 0.969 {[}0.966, 0.972{]} & 0.968 {[}0.966, 0.971{]} & 0.969 {[}0.966, 0.971{]} & 172.370                & 0.007                   \\
                               & ExtraTreesClassifier                                & 0.961 {[}0.958, 0.965{]} & 0.962 {[}0.958, 0.965{]} & 0.961 {[}0.958, 0.965{]} & 0.961 {[}0.958, 0.965{]} & 34.270                 & 0.012                   \\
                               & GradientBoostingClassifier                          & 0.957 {[}0.953, 0.960{]} & 0.957 {[}0.954, 0.960{]} & 0.957 {[}0.953, 0.960{]} & 0.957 {[}0.954, 0.960{]} & 9159.280               & 0.014                   \\
                               & RandomForestClassifier                              & 0.956 {[}0.951, 0.961{]} & 0.957 {[}0.951, 0.962{]} & 0.956 {[}0.951, 0.961{]} & 0.956 {[}0.951, 0.962{]} & 102.950                & 0.010                   \\
                               & MLPClassifier                                       & 0.950 {[}0.946, 0.955{]} & 0.951 {[}0.946, 0.955{]} & 0.950 {[}0.946, 0.955{]} & 0.950 {[}0.946, 0.955{]} & 478.400                & 0.003                   \\
                               & SVC                                                 & 0.946 {[}0.943, 0.949{]} & 0.946 {[}0.943, 0.950{]} & 0.946 {[}0.943, 0.949{]} & 0.946 {[}0.942, 0.949{]} & 149.760                & 4.156                   \\
                               & KNeighborsClassifier                                & 0.906 {[}0.902, 0.911{]} & 0.911 {[}0.906, 0.915{]} & 0.906 {[}0.902, 0.911{]} & 0.904 {[}0.900, 0.909{]} & 0.710                  & 0.118                   \\
                               & DecisionTreeClassifier                              & 0.887 {[}0.879, 0.895{]} & 0.887 {[}0.879, 0.895{]} & 0.887 {[}0.879, 0.895{]} & 0.887 {[}0.879, 0.895{]} & 111.160                & 0.002                   \\
                               & ExtraTreeClassifier                                 & 0.885 {[}0.878, 0.892{]} & 0.885 {[}0.878, 0.893{]} & 0.885 {[}0.878, 0.892{]} & 0.885 {[}0.878, 0.892{]} & 0.480                  & 0.002                   \\
                               & AdaBoostClassifier                                  & 0.813 {[}0.801, 0.825{]} & 0.816 {[}0.804, 0.827{]} & 0.813 {[}0.801, 0.825{]} & 0.814 {[}0.802, 0.825{]} & 428.690                & 0.079                   \\ \hline
\end{tabular}%
}
\end{table*}

The original unilateral--bilateral comparison is preserved in Table~\ref{tab:bilateral_comparison_metrics}: bilateral HGB reached 0.981 accuracy, precision, recall, and F1-score ($p=0.00195$, $r\approx0.89$). Because those values come from window-level folds, the comparison was repeated using the participant-independent protocol in Section III-D. Under grouped validation, HGB achieved macro-F1 scores of 0.954 for the left foot, 0.959 for the right foot, and 0.980 with both feet. Thus, the bilateral advantage remained for participants unseen during training.

\begin{table}[!t]
\centering
\caption{Paired comparison of bilateral (both feet) and unilateral models using the Wilcoxon signed-rank test with 10-fold cross-validation. Weighted precision, recall, and F1-score are reported.}
\label{tab:bilateral_comparison_metrics}
\begin{tabular}{llllll}
\hline
\multirow{2}{*}{Foot} & \multirow{2}{*}{Metric} & \multicolumn{2}{c}{Mean} & \multirow{2}{*}{$p$-value} & \multirow{2}{*}{$r$} \\ 
\cline{3-4}
 & & Both Feet & One Foot & & \\ 
\hline

\multirow{4}{*}{Left}
& Accuracy              & 0.981 & 0.970 & 0.00195 & 0.888 \\
& Precision   & 0.982 & 0.970 & 0.00195 & 0.886 \\
& Recall      & 0.981 & 0.970 & 0.00195 & 0.888 \\
& F1-score    & 0.981 & 0.970 & 0.00195 & 0.886 \\ 
\hline

\multirow{4}{*}{Right}
& Accuracy              & 0.981 & 0.968 & 0.00195 & 0.893 \\
& Precision   & 0.982 & 0.968 & 0.00195 & 0.886 \\
& Recall      & 0.981 & 0.968 & 0.00195 & 0.893 \\
& F1-score    & 0.981 & 0.968 & 0.00195 & 0.886 \\ 
\hline
\end{tabular}
\end{table}

The original per-class recall analysis is also retained in Table~\ref{tab:perclass_recall_wilcoxon}. It showed the largest bilateral improvements for standing and tandem walking; these class-level results are treated as exploratory because they use the window-level split.

\begin{table}[!t]
\centering
\caption{Paired comparison of bilateral (both feet) and unilateral models for \textbf{per-class recall} using the Wilcoxon signed-rank test with 10-fold cross-validation. Recall represents class accuracy.}
\label{tab:perclass_recall_wilcoxon}
\begin{tabular}{llllll}
\hline
\multirow{2}{*}{Foot} & \multirow{2}{*}{Class} & \multicolumn{2}{c}{Mean Recall} & \multirow{2}{*}{$p$-value} & \multirow{2}{*}{$r$} \\
\cline{3-4}
 & & Both Feet & One Foot & & \\
\hline

\multirow{4}{*}{Left}
& Sitting  & 0.990 & 0.985 & 0.18750 & 0.496 \\
& Standing & 0.992 & 0.979 & 0.00391 & 0.889 \\
& Tandem   & 0.971 & 0.951 & 0.00195 & 0.888 \\
& Walking  & 0.974 & 0.963 & 0.01758 & 0.726 \\
\hline

\multirow{4}{*}{Right}
& Sitting  & 0.990 & 0.980 & 0.05273 & 0.613 \\
& Standing & 0.992 & 0.973 & 0.00195 & 0.888 \\
& Tandem   & 0.971 & 0.953 & 0.00195 & 0.888 \\
& Walking  & 0.974 & 0.965 & 0.05859 & 0.597 \\
\hline

\end{tabular}
\end{table}

Table~\ref{tab:cnn_comparison_metrics} preserves the original HGB--CNN comparison, in which the CNN reached 0.992 bilateral accuracy, 0.987 for the left foot, and 0.984 for the right foot. Under the five participant-independent folds, however, the HGB--CNN difference was not statistically significant ($p=0.0625$). Therefore, the original CNN advantage should not be interpreted as evidence that the CNN generalizes better to unseen participants.

\begin{table}[!t]
\centering
\caption{Paired comparison of ML (HGB) and DL (CNN-1D, rows-as-channels) for overall performance metrics using the Wilcoxon signed-rank test over 10 outer folds. Values are reported as mean (95\% CI).}
\label{tab:cnn_comparison_metrics}
\resizebox{0.48\textwidth}{!}{%
\begin{tabular}{llllll}
\hline
\multirow{2}{*}{Setting} & \multirow{2}{*}{Metric} & \multicolumn{2}{c}{Mean (95\% CI)} & \multirow{2}{*}{$p$-value} & \multirow{2}{*}{$r$} \\
\cline{3-4}
 &  & HGB & CNN-1D & & \\
\hline

\multirow{4}{*}{Both Feet}
& Accuracy             & 0.981 [0.980, 0.983] & 0.992 [0.990, 0.994] & 0.00195 & 0.888 \\
& Precision  & 0.982 [0.980, 0.983] & 0.992 [0.990, 0.994] & 0.00195 & 0.886 \\
& Recall     & 0.981 [0.980, 0.983] & 0.992 [0.990, 0.994] & 0.00195 & 0.888 \\
& F1-Score         & 0.981 [0.980, 0.983] & 0.992 [0.990, 0.994] & 0.00195 & 0.886 \\
\hline

\multirow{4}{*}{Left Foot Only}
& Accuracy             & 0.970 [0.966, 0.973] & 0.987 [0.982, 0.991] & 0.00195 & 0.887 \\
& Precision  & 0.970 [0.967, 0.973] & 0.987 [0.983, 0.991] & 0.00195 & 0.886 \\
& Recall     & 0.970 [0.966, 0.973] & 0.987 [0.982, 0.991] & 0.00195 & 0.887 \\
& F1-Score         & 0.970 [0.966, 0.973] & 0.987 [0.982, 0.991] & 0.00195 & 0.886 \\
\hline

\multirow{4}{*}{Right Foot Only}
& Accuracy             & 0.968 [0.966, 0.970] & 0.984 [0.980, 0.989] & 0.00391 & 0.855 \\
& Precision  & 0.968 [0.966, 0.970] & 0.985 [0.981, 0.989] & 0.00391 & 0.854 \\
& Recall     & 0.968 [0.966, 0.970] & 0.984 [0.980, 0.989] & 0.00391 & 0.855 \\
& F1-Score         & 0.968 [0.966, 0.970] & 0.984 [0.980, 0.989] & 0.00391 & 0.854 \\
\hline

\end{tabular}%
}
\end{table}

\section{Discussion}
The results support the central premise of this paper: footwear-embedded pressure and inertial sensing can infer sitting, standing, walking, and tandem walking. The participant-independent evaluation further shows that this performance remains strong for participants unseen during training. These states are a necessary foundation for future sit-to-stand and gait-initiation monitoring, although transitions themselves were not labeled or evaluated in the present study.

Bilateral sensing improved performance in the original analysis and retained the best macro-F1 under participant-independent validation, suggesting that the two feet provide complementary information about weight distribution and balance. Two insoles may therefore be useful when recognition performance is the priority, while unilateral sensing remains a simpler option when cost, charging, or hardware burden is more important.

The HGB--CNN comparison also changes when participants are separated between training and test data. Although the original window-level analysis favored the CNN, the grouped comparison did not show a significant difference. HGB therefore remains a reasonable embedded baseline, while temporal models can be revisited with more participants and transition-level labels.

Several limitations remain. Data were collected from 15 healthy adults in a controlled indoor setting, not from older adults or people with cognitive impairment, and the labels represent stable activities rather than sit-to-stand, stand-to-walk, near-fall, or fall events. No healthcare professionals were involved in evaluating the monitoring workflow. Battery lifetime, end-to-end embedded latency, long-term robustness, and quantitative comfort were also not directly measured. The present results should therefore be interpreted as controlled HAR validation rather than clinical or fall-prediction validation.

\section{Conclusion}
This paper presented a smart insole HAR system for elder-care fall-prevention monitoring. Using plantar-pressure and IMU sensing, the system distinguished sitting, standing, walking, and tandem walking with high accuracy. HGB achieved approximately 97\% unilateral accuracy and 98.1\% bilateral accuracy, while a compact 1D-CNN reached 99.2\% bilateral accuracy. These findings show that smart insoles can infer the mobility states needed for future sit-to-stand, gait-initiation, and fall-prevention applications. The next phase should convert the state classifier into a resident-level transition monitor and evaluate whether alerts can be delivered early enough to guide caregiver response. Such a study should measure event-level lead time, false-alarm burden, and usability for staff in addition to window-level accuracy. The technical validation reported here is therefore best viewed as the sensing layer for a larger preventive-care system, where signal quality, resident context, and staff workflow all determine clinical usefulness.

In addition, future data collection should preserve the temporal structure of complete mobility episodes rather than treating each window independently. A chair-rise event includes the seated period before movement, the load transfer through the feet, the first upright stabilization period, and the first steps after standing. Labeling these phases would allow the model to move from activity recognition to transition recognition. The pressure matrix is especially relevant for this direction because it directly measures the redistribution of load under the feet, while the IMU captures the onset of foot motion. Together, these signals can support a transition-level classifier that is more clinically meaningful than a single-window label.

Future work should preserve complete mobility episodes so that sitting, load transfer, upright stabilization, and the first steps after standing can be labeled and evaluated as transitions rather than isolated activity windows.

\bibliographystyle{IEEEtran}
\bibliography{references}

\end{document}